\documentclass[runningheads]{llncs}

\usepackage{eccv}

\usepackage{eccvabbrv}

\usepackage{graphicx}
\usepackage{booktabs}

\usepackage{colortbl}
\usepackage{multirow}
\usepackage{amsmath}
\usepackage{pifont}

\usepackage{subcaption}

\newcommand{\GR}[0]{\cellcolor[gray]{0.9}}
\newlength\savewidth\newcommand\shline{\noalign{\global\savewidth\arrayrulewidth
  \global\arrayrulewidth 1pt}\hline\noalign{\global\arrayrulewidth\savewidth}}

\usepackage[accsupp]{axessibility}  % Improves PDF readability for those with disabilities.

\usepackage{hyperref}

\usepackage{orcidlink}

\begin{document}

\title{Aligning Human Sense: Calibrated Distributional Reward Learning for Video Generation}

\newcommand{\affmark}[1]{\textsuperscript{#1}}
\newcommand{\emailmark}{\textsuperscript{\ensuremath{\dagger}}}
\newcommand{\cormark}{\textsuperscript{*}}

\author{
Nai-Xin Zhai\affmark{1,11}\emailmark \and
Weihua Cheng\affmark{2,11} \and
Dexu Yu\affmark{3} \and
Yikai Gu\affmark{4} \and
Hanwen Du\affmark{5,11} \and
Junchen Fu\affmark{6,11} \and
Chenxi Huang\affmark{7} \and
Yingwei Song\affmark{8} \and
Liyuan Lillian Ma\affmark{9} \and \\
Yang Ran\affmark{3} \and
Youhua Li\affmark{1}\cormark \and
Yongxin Ni\affmark{10}\cormark
}

\authorrunning{N. Zhai et al.}
\titlerunning{Calibrated Distributional Reward Learning for Video Generation}

\institute{
\parbox{0.98\textwidth}{
\centering
\scriptsize
\setlength{\baselineskip}{8.2pt}
\affmark{1} City University of Hong Kong \quad
\affmark{2} ShanghaiTech University \quad
\affmark{3} Fenz.AI 
\\[-0.15em]
\affmark{4} University of California, Berkeley \quad
\affmark{5} The Ohio State University 
\\[-0.15em]
\affmark{6} University of Glasgow \quad
\affmark{7} Columbia University \quad
\affmark{8} University of Arizona
\\[-0.15em]
\affmark{9} GMI Cloud \quad
\affmark{10} National University of Singapore \quad
\affmark{11} DeciLix Lab
\\
\emailmark~Email: \texttt{nancyzhzhai@gmail.com}
\\
\cormark~Co-corresponding: 
\texttt{youhuali2-c@my.cityu.edu.hk}; 
\texttt{niyongxin@u.nus.edu}
}
}

\maketitle

\begin{abstract}

Video generation is central to AI-powered content creation. 
The alignment with human preferences is one of the key metrics for measuring the quality of the generated videos. Despite significant progress in visual quality, three key challenges remain: 
1) The reliability of reward signals is constrained by the quality of human preference data, which is often corrupted by subjective noise and bias. 
2) Standard scalar reward models collapse multi-aspect human preference into a single value, leading to the loss of dynamic trade-offs across multiple dimensions of human preference. 
3) In policy optimization, the widely adopted KL-divergence imposes only local constraints, failing to capture holistic human preference. 
To address these challenges, we propose a unified, preference-aware learning framework for video generation. 
First, we propose elite-guided filtering to calibrate preference data and construct reliable supervision for reward-model training.
We then model video quality as a multidimensional reward distribution to capture the uncertain nature of human preference, and use the Wasserstein distance to align it with the empirical human preference distribution. 
Finally, we introduce Wasserstein-based distributional alignment in GRPO, guiding the policy’s video generation to match the global structure of human video preference. 
Experiments on reward modeling and video generation show that our approach improves the reliability of reward signals and the perceptual consistency of generated videos. 
Our code is available at \url{https://github.com/alignhs26/ahs}.
\end{abstract}

\section{Introduction}
\label{intro}

Video generation has become a core technology in AI-driven content creation, with growing applications in film and television production, virtual interaction, and digital simulation. 
Recent methodological advances in deep generative modeling have substantially improved the temporal coherence and perceptual fidelity of synthesized videos~\cite{ho2020denoising, lipman2022flow, zhang2025packing}. 
This progress is accelerating adoption in creative and interactive applications~\cite{meng2025echomimicv2, ni2025content, fu2026llmpopcorn}. 
However, as baseline generation quality approaches basic plausibility requirements, further performance gains increasingly depend on alignment with human preferences.
Given the inherently multidimensional and context-sensitive nature of human judgment, achieving effective preference alignment has emerged as a critical bottleneck that determines the quality of generated video content.

To address this challenge, human feedback-based alignment frameworks developed initially for language models, including reinforcement learning from human feedback (RLHF) and direct preference optimization (DPO), have been adapted to video generation~\cite{wu2025densedpo, meng2025identity}.
Recent work has made several advancements to particularly address the inherent variability and expressive richness of human preferences, which often impede consistent recovery of underlying intent.
One line of work enhances the granularity and structure of human feedback in annotation data, translating vague, holistic preference into structured and traceable supervisory signals to enable finer-grained and more transparent alignment~\cite{wu2025densedpo, meng2025identity, wang2025vr}. 
Another line aims to reduce or eliminate manual annotation by either bootstrapping reward models from limited human labels~\cite{wang2025vr} or reframing preference learning as a generative process, enabling scalable and evolvable preference acquisition~\cite{wu2025rewarddance}.

\begin{figure*}[t]
  \centering
  \includegraphics[width=\textwidth]{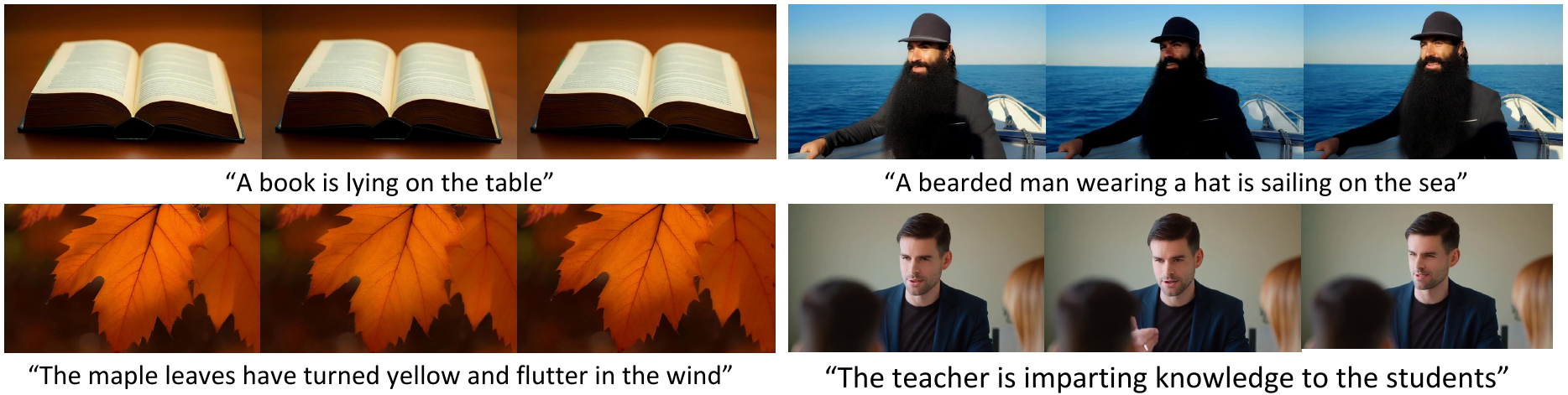}
  \caption{Our framework produces high-quality text-to-video generation that closely aligns with human preferences, demonstrating significant improvements in both perceptual plausibility and semantic fidelity. 
  }
  \label{fig:showcase}
\end{figure*}

Although existing studies have explored various strategies to improve alignment, they continue to face three key issues that prevent genuine human alignment:
1) The reliability of the reward signal is limited by the quality of human preference data, which is often affected by subjective noise and bias, leading to deviation from the accurate preference distribution~\cite{wang2020reinforcement}.  
2) Existing reward models~\cite{xu2024visionreward,he2024videoscore} typically assume that video quality can be represented by a single scalar value and optimize it using objective functions such as pairwise ranking or binary classification. In practice, human preferences arise from multiple interrelated dimensions, including but not limited to semantic plausibility, visual naturalness, and temporal coherence. These dimensions exhibit complex couplings and trade-offs, and forcing them into a single scalar not only discards essential structural information but also frequently results in inconsistent preference representations~\cite{he2024videoscore,xu2024visionreward}. 
3) Additionally, the KL divergence objective used in RLHF and GRPO~\cite{shao2024deepseekmath,xue2025dancegrpo,liu2026flow} applies constraints only locally within each denoising step of the diffusion process, functioning as a pointwise and instantaneous mechanism for distributional alignment. It does not account for temporal dependencies or semantic evolution across steps, and therefore cannot reflect the holistic human preference for video. To avoid penalties from local distributional deviations, the policy tends to select low-risk denoising trajectories, often resulting in motion-conservative video outputs that fail to capture complex dynamic interactions. Although such samples may match the reference behavior in local statistical properties, they consistently diverge from human preference in terms of global visual plausibility and temporal coherence. 

To address the alignment discrepancy stemming from the three issues discussed above, we propose a unified framework for human-aligned video preference learning that systematically tackles their root causes. Specifically, as shown in Fig.~\ref{fig:pipeline}, our approach proceeds in three stages:
1)~\textbf{Elite-Guided Preference  Calibration:} To extract high-quality and reliable supervision signals from noisy and biased human annotations, we avoid treating all samples equally. Instead, we quantify the intrinsic reliability of each sample using consistency metrics.
High-confidence samples identified through this process are used to train an initial scoring model, which is then employed as a scorer to re-evaluate the quality of low-confidence data and recalibrate the overall preference distribution, thereby providing higher-quality training signals.
2)~\textbf{Preference-Aware Reward Modeling:} To address the limitation of reward models that rely on a single scalar and thus fail to capture the multidimensional couplings within human preference, we model video quality as a multidimensional reward distribution defined over multiple perceptual and semantic dimensions rather than a single value. 
Based on this formulation, we employ the Wasserstein distance as a distribution-level alignment objective to align the model-predicted reward distribution with the empirical distribution of human preference, ensuring that the predicted rewards more faithfully reflect the diversity of real human preference.
3)~\textbf{Preference-Guided Policy Optimization:} We further extend the concept of distributional alignment to the policy optimization stage by introducing the Wasserstein distance as an alternative to the KL divergence within the single-step update framework of Group Relative Policy Optimization (GRPO)~\cite{shao2024deepseekmath}.
In this stage, the policy governs the generation of video sequences, and the human video-preference distribution serves as the reference for alignment.
During each policy update, we align the global characteristics of the generated video distribution rather than merely matching local statistical densities.
The Wasserstein distance acts as a robust behavioral alignment metric, guiding the policy's output distribution towards the empirical distribution of human video preferences.

Our contribution can be summarised as follows:
\begin{enumerate}
\item We propose a unified framework for human-aligned video preference learning that systematically addresses the misalignment across three levels: data reliability, reward representation, and policy alignment.
\item We propose an elite-guided calibration framework to evaluate annotation reliability. Building on this, we leverage this representation to achieve alignment in both reward modeling and policy optimization respectively.
\item We conduct comprehensive experiments on reward modeling and video generation across multiple metrics, verifying the effectiveness of our framework in aligning the learned reward distribution with human preference.
\end{enumerate}
\section{Related Work}
\label{related_work}
\subsubsection{Data Calibration}
Human preference data underpin the alignment of large language models (LLMs) and video generation systems; however, its scalability is limited by the cost and inconsistency of human annotations. 
Recent studies address these challenges through data calibration, emphasizing both the selective use of high-quality data and the correction of noise in low-quality supervision. 
Automated surrogates, such as reward models (e.g., the helpful–harmless objective~\cite{bai2022helpfulharmless}) and lightweight alignment strategies~\cite{zhou2023lima}, reduce reliance on costly human feedback by capturing alignment signals at scale. 
Meanwhile, data-centric approaches focus on quality-based selection and reweighting. 
DEITA~\cite{liu2024deita}, which quantifies complexity, quality, and diversity to identify high-impact samples, shows that 6K curated examples can outperform ten times more random data. 
Complementary studies on filtering, clustering, and IFD-based self-calibration~\cite{cheng2024unleashingpipeline,qin2024datasurvey,li2024quantity} further show that alignment gains mainly come from data quality and calibration, rather than data volume.

\subsubsection{Human Feedback Alignment} 
Recent studies in visual generation emphasize the collection of human preference data, training reward models, and optimization of generative models.
VideoScore~\cite{he2024videoscore} builds a large multi-dimensional human rating dataset and an automatic scorer to simulate feedback. VisionReward~\cite{xu2024visionreward} models fine-grained, interpretable preferences such as composition, motion, and prompt consistency for better alignment. VideoAlign~\cite{liu2025humanfeedback} directly applies preference optimization to video generation, while UnifiedReward~\cite{wang2025unified} proposes a cross-task, cross-modal reward model for broader generalization.
Overall, the field is moving toward fine-grained, multi-dimensional, and unified human preference alignment, though challenges remain in data scale and reward-model generalization.
\subsubsection{Video Generation}
Video generation has advanced markedly, driven by diffusion models~\cite{ho2020denoising, rombach2022high,blattmann2023alignlatents, singer2022make} and flow-matching models~\cite{lipman2022flow, liu2022flow,kong2024videocrafter2, zhang2024kling15}.
Despite steady gains in resolution and stability, generated results often remain misaligned with human preferences, manifesting as semantic misinterpretation, visual content deviating from intent descriptions, and mismatches between perceived quality and user judgments.
To mitigate this alignment gap, human feedback-based approaches have been transferred from language to visual generation.
Methods based on DPO~\cite{liu2024videodpo,wu2025densedpo} apply Direct Preference Optimization to align both visual quality and semantic relevance. 
Methods based on GRPO~\cite{xue2025dancegrpo,liu2026flow} adopt Group Relative Policy Optimization across diffusion and flow-matching models, unifying reward models and stabilizing policy optimization for aligning human preferences in video generation. 
Regardless of the optimization algorithm, the pivotal factor remains the quality and distribution of human preference data, because aligning the model output to human subjective standards ultimately depends on capturing and modelling those preferences accurately.
\section{Method}
\label{sec:method}

\begin{figure*}[t]
  \centering
  \includegraphics[width=\textwidth]{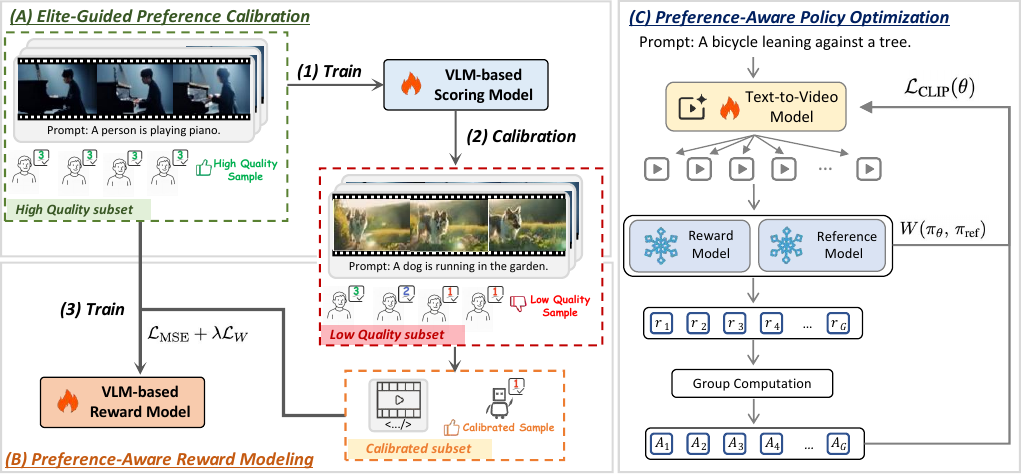}
    \caption{Overview of our framework. (A) Elite-Guided Preference Calibration trains a scoring model on high-consistency annotations and calibrates the low-quality subset. (B) Preference-Aware Reward Modeling fine-tunes a VLM-based reward model with an MSE loss and a Wasserstein alignment loss. (C) Preference-Guided Policy Optimization applies Wasserstein-based GRPO to align group-wise video generation with the reference reward model.}
  \label{fig:pipeline}
\end{figure*}

Our method proceeds in three stages (Fig.~\ref{fig:pipeline}). 
Elite-Guided Preference Calibration (Sec.~\ref{sec:elite-guided-calibration}) first partitions annotations by intrinsic agreement, trains a scorer on the high-consistency subset, and calibrates low-consistency samples to expand a high-confidence training set. 
Preference-Aware Reward Modeling (Sec.~\ref{sec:reward-modeling}) then fits a VLM-based reward model on the calibrated data. 
Preference-Guided Policy Optimization (Sec.~\ref{sec:policy-alignment}) finally fine-tunes the video generator with Wasserstein-based GRPO, using the trained reward model to improve perceptual quality and text–video alignment.

\subsection{Elite-Guided Preference Calibration}
\label{sec:elite-guided-calibration}

To extract high-quality and reliable supervision signals from noisy and biased human annotations, we propose the Elite-Guided Preference Calibration (EGPC) framework. 
The core idea is to treat human annotations with high inter-annotator consistency as elite signals to construct a reliable scoring proxy, thereby enabling systematic calibration of the preference distribution at the full-dataset scale. 
The procedure consists of three stages:

\subsubsection{High-Quality Subset Selection}

Given the original annotation dataset $\mathcal{D} = \{(v_i, p_i, \mathbf{x}_i)\}_{i=1}^N$,
where $v_i$ denotes a generated video, $p_i$ represents the corresponding prompt, and 
$\mathbf{x}_i = \{x_{i1}, \dots, x_{in}\}$  is the independent ratings provided by $n$ human annotators 
on a discrete rating scale 
$\mathcal{S} = \{s_1, \dots, s_k\}$.

We define a high-quality subset $\mathcal{D}_H$ based on annotation consistency. 
This subset aims to capture stable and reproducible human judgments, thereby providing trustworthy supervision for downstream modeling. 
Specifically, a sample $i$ is included in $\mathcal{D}_H$ if it meets a standard of high consistency: that is, all annotators assign identical scores, or, if a discrepancy exists, there is at most a single outlier rating $x_{ij}$ which satisfies the condition that its absolute difference from the mode of the ratings does not exceed 1, i.e., $|x_{ij} - \text{mode}(\mathbf{x}_i)| \leq 1$. Here, $\text{mode}(\mathbf{x}_i)$ denotes the most frequently occurring rating among the annotators for sample $i$. 
All remaining samples that do not meet this high consistency standard constitute the low-quality subset $\mathcal{D}_L = \mathcal{D} \setminus \mathcal{D}_H$. 
This selection process ensures that the samples in $\mathcal{D}_H$ possess a high degree of annotation consistency and reliability, thereby providing a more robust and trustworthy supervisory signal.

\subsubsection{Elite Scorer Training}
Using $\mathcal{D}_H$, we train a parameterized scoring function $f_{\theta}: (v, p) \mapsto \hat{x} \in \mathbb{R}$ by minimizing the empirical risk between predictions and human ratings:  
\begin{equation}
\theta^* = \arg\min_{\theta} \frac{1}{|\mathcal{D}_H|} \sum_{(v_i, p_i, \mathbf{x}_i) \in \mathcal{D}_H} \ell\big(f_{\theta}(v_i, p_i), \,\text{agg}(\mathbf{x}_i)\big),
\end{equation}
where $\text{agg}(\cdot)$ is an aggregation operator, and $\ell$ is a regression loss. The resulting model $f_{\theta^*}$, referred to as the Elite Scorer, aims to approximate the mapping underlying expert-consensus judgments.
\subsubsection{Full-Set Preference Calibration}
The trained elite scorer is applied to the low-quality subset $\mathcal{D}_L$ to obtain calibrated scores: for each $(v_i, p_i, \mathbf{x}_i) \in \mathcal{D}_L$, compute $\hat{x}_i = f_{\theta^*}(v_i, p_i)$, and construct the calibrated subset $\mathcal{D}_L^* = \{(v_i, p_i, \hat{x}_i) \mid (v_i, p_i, \mathbf{x}_i) \in \mathcal{D}_L\}$. Finally, we merge the original high-quality samples with the calibrated ones to form an augmented high-quality supervision set:  
\begin{equation}
\mathcal{D}_H^{\text{aug}} = \mathcal{D}_H \cup \mathcal{D}_L^*.
\end{equation}  
This augmented set preserves the structural integrity of expert consensus while substantially expanding the scale of effective supervision, thus providing a consistent preference signal foundation for subsequent reward modeling.
\subsection{Preference-Aware Reward Modeling}
\label{sec:reward-modeling}

To overcome the limitation of single-scalar rewards that fail to capture the complexity of multidimensional human preferences, we model video quality as a probability distribution over a $K$-dimensional score space rather than a single value.
For each video--prompt pair $(v_i,p_i)$, let $\mathbf{y}_i\in\mathbb{R}^{K}$ denote its calibrated multidimensional human-preference label.
The empirical target distribution is represented as $\delta_{\mathbf{y}_i}$.

For a reward model parameterized by $\theta$, we predict a conditional distribution $p_{\theta}(\mathbf{z}\mid v_i,p_i)$, where $\mathbf{z}\in\mathbb{R}^{K}$ denotes a possible multidimensional preference score.
To ensure consistency between the predicted and empirical distributions, we introduce Wasserstein distance~\cite{feydy2019geomloss} as a complementary alignment objective.
For a given sample $(v_i,p_i,\mathbf{y}_i)$, the distribution alignment loss is defined as:
\begin{equation}
\mathcal{L}_{W}(v_i,p_i)
=
W_p\big(p_{\theta}(\mathbf{z}\mid v_i,p_i),\, \delta_{\mathbf{y}_i}\big).
\end{equation}

The total training objective combines conventional regression supervision with this distributional alignment:
\begin{equation}
\mathcal{L}_{\mathrm{total}}
=
\mathcal{L}_{\mathrm{MSE}} + \lambda\mathcal{L}_{W},
\end{equation}
where $\lambda>0$ controls the relative importance of the Wasserstein alignment term. The mean squared error (MSE) term is defined as:
\begin{equation}
\mathcal{L}_{\mathrm{MSE}}
=
\mathbb{E}_{(v,p,\mathbf{y})\sim\mathcal{D}}
\left\| f_{\theta}(v,p)-\mathbf{y} \right\|_2^{2},
\end{equation}
where $f_{\theta}(v,p)$ is the predicted multidimensional reward vector.

\subsection{Preference-Guided Policy Optimization}
\label{sec:policy-alignment}

To overcome the limitation of KL divergence in failing to capture holistic human preferences, we extend the concept of distributional alignment to the policy optimization stage by proposing Wasserstein-based Group Relative Policy Optimization (WGRPO). 
Specifically, we replace the KL divergence with the Wasserstein distance within the single-step update framework of GRPO. 
Unlike the KL divergence, which only matches local statistical densities, the Wasserstein distance aligns the global characteristics of the generated video distribution, serving as a behavioral alignment constraint that guides the policy's overall output to progressively approximate human preference.

Specifically, given a prompt $\mathbf{c}$, the old policy $\pi_{\theta_{\text{old}}}$ first samples a group of $G$ candidate sequences $\{\mathbf{o}_1, \dots, \mathbf{o}_G\}$ following the group sampling paradigm of DeepSeek-R1~\cite{deepseekr1}, where each candidate $\mathbf{o}_i$ corresponds to a trajectory consisting of state–action pairs $\{(s_{t,i}, a_{t,i})\}_{t=1}^{T}$. For each timestep, we compute the importance ratio between the new and old policies as 
\begin{equation}
\rho_{t,i}=\frac{\pi_\theta(a_{t,i}|s_{t,i})}{\pi_{\theta_{\text{old}}}(a_{t,i}|s_{t,i})}.  
\end{equation}
The reward $r_i$ obtained for each trajectory is normalized within the group to form a group-relative advantage,
\begin{equation}
\label{eq:wgrpo-adv}
A_i = \frac{r_i - \text{mean}({r_1,\dots,r_G})}{\text{std}({r_1,\dots,r_G})},
\end{equation}
which stabilizes optimization, reduces reward-scale variance, and preserves the relative preference relationships among candidates. Building upon the PPO framework~\cite{schulman2017ppo}, the surrogate objective for one timestep $(t, i)$ is defined as
\begin{equation}
\label{eq:wgrpo_timestep_loss}
L_{t,i}(\theta)
= \min\bigl( \rho_{t,i}\, A_i,\; \operatorname{clip}(\rho_{t,i},\, 1-\epsilon,\, 1+\epsilon)\, A_i \bigr),
\end{equation}
where the clipping term limits the update magnitude and prevents excessively large policy shifts. Averaging over all timesteps of trajectory $\mathbf{o}_i$ yields the per-sample objective,
\begin{equation}
\label{eq:wgrpo_sample_loss}
\mathcal{L}_i(\theta) = \frac{1}{T}\sum\nolimits_{t=1}^{T} L_{t,i}(\theta),
\end{equation}
and the final group-level objective is obtained by taking the expectation over all groups sampled from the old policy,
\begin{equation}
\label{eq:wgrpo_clip_final}
\mathcal{L}_{\text{CLIP}}(\theta)
= \mathbb{E}_{\substack{ \{\mathbf{o}_i\} \sim \pi_{\theta_{\text{old}}} \\ \{a_{t,i}\} \sim \pi_{\theta_{\text{old}}} }}
\left[ \frac{1}{G}\sum\nolimits_{i=1}^{G} \mathcal{L}_i(\theta) \right].
\end{equation}

To further align the global behavioral characteristics of the new policy with human-preferred generation patterns, we incorporate a distributional alignment term based on the Wasserstein distance between the output distribution $\pi_\theta$ and the reference policy $\pi_{\text{ref}}$. The Wasserstein term measures the minimal transport cost required to transform one output distribution into another under a ground metric $d$ on the output space $\mathcal{Y}$, serving as a behavioral alignment constraint that captures global semantic and perceptual consistency. 
Combining the clipped GRPO surrogate with a Wasserstein alignment penalty yields the following minimization objective:
\begin{equation}
\mathcal{L}_{\mathrm{WGRPO}}(\theta)
= -\mathcal{L}_{\text{CLIP}}(\theta)
+ \beta\, W\!\left(\pi_\theta,\, \pi_{\text{ref}}\right),
\end{equation}
where $\beta>0$ controls the strength of the distributional alignment penalty.
 \section{Experiments}
\label{sec:experiments}

\begin{table*}[t]
    \centering
        \caption{Comparison of sub-dimension scores before and after alignment for Wan, CogVideo and ModelScope. Fine-tuning with our two-dimensional calibrated reward models shows stronger human-preference alignment in generated videos. RM Type denotes Reward Model Type.}
    \renewcommand{\arraystretch}{1.2}
\resizebox{\textwidth}{!}{
    \begin{tabular}{cc|cccccccc}
    \hline
    \multicolumn{1}{c}{\textbf{Model}} & \multicolumn{1}{c}{\textbf{\begin{tabular}[c]{@{}c@{}}RM\\ Type\end{tabular}}} & \multicolumn{1}{c}{\textbf{\begin{tabular}[c]{@{}c@{}}Motion\\ smooth.\end{tabular}}} & \multicolumn{1}{c}{\textbf{\begin{tabular}[c]{@{}c@{}}Dynamic\\ degree\end{tabular}}} & \multicolumn{1}{c}{\textbf{\begin{tabular}[c]{@{}c@{}}Spatial\\ relation\end{tabular}}} & \multicolumn{1}{c}{\textbf{Scene}} & \multicolumn{1}{c}{\textbf{\begin{tabular}[c]{@{}c@{}}Appear.\\ style\end{tabular}}} & \multicolumn{1}{c}{\textbf{\begin{tabular}[c]{@{}c@{}}Subject\\ consist.\end{tabular}}} & \multicolumn{1}{c}{\textbf{\begin{tabular}[c]{@{}c@{}}Back.\\ consist.\end{tabular}}} & \textbf{Avg.} \\ \shline
    \multirow{2}{*}{\begin{tabular}[c]{@{}c@{}}Wan2.1\\ T2V-1.3B\end{tabular}} & Raw (Full) & 88.25 & 35.58  & 61.23 & 69.65 & 42.21 & 88.89 &  91.32 & 68.16 \\
     & \GR Calibrated  & \GR 89.32  & \GR 36.48 & \GR 63.23 & \GR 71.32 & \GR 42.25 & \GR 91.90 & \GR 92.55 & \GR \textbf{69.58} \\ \cline{1-10}
     \multirow{2}{*}{
     \begin{tabular}[c]{@{}c@{}}CogVideoX\\ 2B\end{tabular}} & Raw (Full)  & 86.53 & 31.33 & 54.52 & 42.15 & 75.58 & 87.35 & 87.32 & 66.39 \\
     & \GR Calibrated & \GR 90.25 & \GR 31.99 & \GR 56.89  & \GR 47.15 & \GR 76.95 & \GR 86.45 & \GR 88.57 &  \GR \textbf{68.32} \\ \cline{1-10}
     \multirow{2}{*}{ModelScope} & Raw (Full) & 81.53 & 27.77 & 59.33 & 40.39 & 70.58 & 82.12 & 85.33 & 63.86 \\
     & \GR Calibrated & \GR 82.34 & \GR 29.54 & \GR 63.12 & \GR 45.56 & \GR 73.98 & \GR  80.25 & \GR 86.66 & \GR \textbf{65.92} \\
    \hline
    \end{tabular}
    }
    \label{tab:main_result_videogen}
\end{table*}

\subsection{Dataset}
This experiment employs two publicly available benchmark datasets: VidProM~\cite{wang2024vidprom} for text-conditional video generation, and Video-Bench~\cite{han2025video} for reward modeling targeting human alignment.

\subsubsection{Video-Bench}
Video-Bench~\cite{han2025video} provides a collection of annotated video–prompt pairs, supporting multi-dimensional quality assessment. The dataset comprises 8799 unique video samples, generated by four open-source and three commercial text-to-video models. Each sample is rated independently by four annotators across nine dimensions, specifically:  
\begin{itemize}
    \item Five alignment dimensions (e.g., Action, Color Consistency), assessing semantic fidelity between generated content and the textual instruction;
    \item Four video quality dimensions (e.g., Motion Effects, Aesthetic Quality), evaluating perceptual quality and physical plausibility.
\end{itemize}

\subsubsection{VidProM}
For policy training, we curate 50,000 text prompts from VidProM~\cite{wang2024vidprom}, a collection of community-generated prompts from the official Pika Labs Discord channel from October 2023 to March 2024. The selected prompts cover diverse video-generation scenarios, actions, and visual styles, providing varied text conditions for policy learning.

\subsection{Experimental Setting}

\subsubsection{Dataset Calibration}
Following the procedure in Sec.~\ref{sec:elite-guided-calibration}, we first isolate a high-consistency subset from Video-Bench. 
The subset is selected via inter-annotator agreement and annotation stability and serves as the foundation for training an elite scoring model. Using Qwen2-VL-2B~\cite{wang2024qwen2} as the backbone, this model learns to predict video quality aligned with human perceptual judgments. 
Subsequently, we reweight the remaining samples by the model's confidence and perform hard filtering to discard low-scoring outliers, thereby suppressing tail-distributed noise and yielding a calibrated training distribution.

\subsubsection{Reward Modeling}
The reward model is implemented by fine-tuning Qwen2-VL-2B~\cite{wang2024qwen2} with LoRA (rank $=8$)~\cite{hu2022lora}, batch size 32, and learning rate $2\times10^{-6}$. 
All other hyperparameters follow VideoAlign~\cite{liu2025humanfeedback}.
Consistent with practices in previous work~\cite{liu2026flow, he2024videoscore}, we evaluate on a held-out set of 100 high-fidelity video pairs, reporting pairwise accuracy both with ties and excluding ties~\cite{deutsch2023ties}.

\begin{table}[t]
    \small
    \centering
        \caption{Comparison of Reward Models on the pairwise accuracy on the Action and Motion Effects dimensions under two data scales (Equal and Full).}
    \setlength{\tabcolsep}{5pt}
    \renewcommand{\arraystretch}{1.2}
    \begin{tabular}{c|cccc}
    \hline
     \multirow{2}{*}{\textbf{Scale}} & \multicolumn{2}{c}{\textbf{Action}}    & \multicolumn{2}{c}{\textbf{Motion Effects}} \\  \cline{2-5}
                   & \textbf{w/ Ties}       & \textbf{w/o Ties}      & \textbf{w/ Ties}          & \textbf{w/o Ties}        \\  \shline
    Raw(Equal)     & 54.65         & 74.24         & 67.68             & 80.30         \\ \rowcolor[gray]{.9}
    Curated(Equal) & \textbf{59.60} & \textbf{78.79} & \textbf{68.69}    & \textbf{84.85}   \\ \shline
    Raw(Full)      & 58.59          & 80.30          & 65.66             & 86.36           \\ \rowcolor[gray]{.9}
    Calibrated     & \textbf{60.61} & \textbf{83.33} & \textbf{71.72}    & \textbf{87.88}   \\
    \hline
    \end{tabular}
    \label{tab:main_result_rm}
\end{table}

\subsubsection{Preference-Guided Model Adaptation}
\label{subsec:adaptation}
We adapt three open-source text-to-video base models, Wan2.1-T2V-1.3B~\cite{wan2025wan}, CogVideoX-2B~\cite{yangcogvideox}, and ModelScope-T2V~\cite{wang2023modelscope}, using reward signals from the preference-aware reward model trained in Sec.~\ref{sec:reward-modeling}.
Training is conducted within the Video-Bench framework, ensuring fair comparison and reproducibility. 
We refer to this stage as preference-guided policy adaptation because it applies GRPO-style updates to align generated videos with calibrated human-preference rewards.
To mitigate prompt leakage, we use GPT-4o to rewrite and extend the original test prompts. 
Each original prompt is paraphrased into three stylistically and semantically distinct variants, yielding a more robust and generalizable evaluation set.

\subsubsection{Further Details}
\label{exp_details}
All experiments are conducted on 8 NVIDIA H100 GPUs. For policy optimization, we use a sampling timestep $T = 20$ and a group size $G = 16$. We use LoRA with $\alpha = 64$ and $r = 32$.

\subsection{Main Results}
\subsubsection{Reward Modeling Performance}
Table~\ref{tab:main_result_rm} reports pairwise accuracy for the reward model on Action Consistency and Motion Effectiveness. 
Training on curated calibrated data improves accuracy over training on the raw data, showing the benefit of calibration. 
With the full dataset, the model reaches 60.61 with ties and 83.33 without ties on Action Consistency, and 71.72 with ties and 87.88 without ties on Motion Effectiveness. 
These gains indicate that calibration helps the reward model evaluate both action consistency and motion effectiveness. 
As illustrated in Fig.~\ref{fig:showcase}, our method achieves superior alignment with high-quality human preference, yielding outputs with high perceptual fidelity and overall performance improvements.

\subsubsection{Video Generation Quality Improvement}
Table~\ref{tab:main_result_videogen} compares three open-source video generation models before and after optimization with our calibrated reward model. 
The optimized models improve mainly on motion-related metrics, including Motion Smoothness and Dynamic Degree. 
For example, CogVideoX improves from 86.53 to 90.25 in Motion Smoothness and from 31.33 to 31.99 in Dynamic Degree. 
The total composite score also increases by 2.1\% for Wan2.1, 2.9\% for CogVideoX, and 3.2\% for ModelScope. 
The main exception is a small drop in Subject Consistency, suggesting a trade-off from selecting samples that favor dynamic quality over static content consistency.

\begin{figure*}[t]
  \centering
  \includegraphics[width=\textwidth]{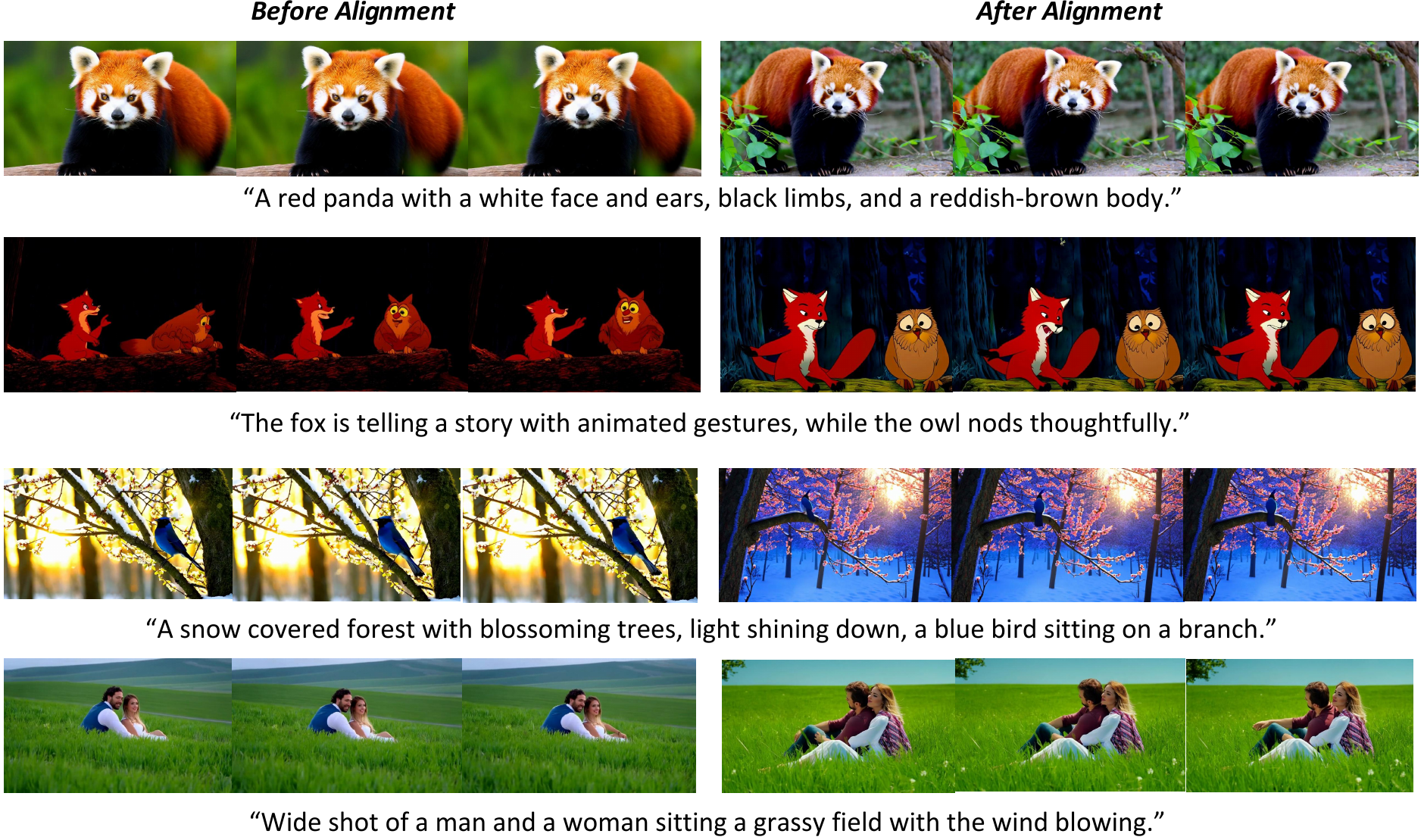}
  \caption{
  Alignment effect comparison: The left column displays results from models before alignment, while the results after our alignment are shown on the right.}
  \label{fig:align_showcase}
\end{figure*}

\subsection{Ablation Study} 

To validate the effectiveness of $\mathcal{L}_{\text{W}}$, we conduct a two-part ablation study focusing on both Reward Modeling and Policy Optimization. The results in Tab.~\ref{tab:abla_rm_wass_loss} and Tab.~\ref{tab:abla_result_videogen} demonstrate the benefits of choosing the Wasserstein Distance.
\begin{table}
    \centering
        \caption{Ablation study on Wasserstein distance loss for reward modeling.}
    \setlength{\tabcolsep}{8.5pt}
    \renewcommand{\arraystretch}{1.1}
    \begin{tabular}{c|cccc}
    
    \hline
     \multirow{2}{*}{$\mathcal{L}_{\text{W}}$} & \multicolumn{2}{c}{Action} & \multicolumn{2}{c}{Motion Effects} \\ \cline{2-5}
      & w/ Ties & w/o Ties & w/ Ties & w/o Ties \\ \shline
    \ding{55} & 58.98 & 81.25 & 68.33 & 86.11 \\ \rowcolor[gray]{.9}
    \ding{51} & \textbf{60.61 }& \textbf{83.33} & \textbf{71.72} & \textbf{87.88} \\      
    \hline
    \end{tabular}
    \label{tab:abla_rm_wass_loss}
\end{table}

\subsubsection{Reward Modeling Performance}
Table~\ref{tab:abla_rm_wass_loss} shows the effect of $\mathcal{L}_{\text{W}}$ on the reward model's pairwise accuracy for Action Consistency and Motion Effectiveness. 
Adding $\mathcal{L}_{\text{W}}$ improves all reported metrics. For Motion Effectiveness, accuracy increases from 68.33 to 71.72, a gain of 3.39. 
For Action Consistency, accuracy improves from 58.98 to 60.61. 
Wasserstein distance provides a distribution-level constraint by measuring the transport cost between the predicted reward distribution and the human-preference reference distribution, rather than only penalizing local deviations. 
This property is well suited to video preferences that involve global semantic consistency, temporal coherence, and multi-dimensional quality attributes, helping the reward model better distinguish quality differences in complex dynamic content.

\begin{table*}[t]
    \centering
        \caption{Ablation study on Wasserstein distance constraint for video generation.}
    \renewcommand{\arraystretch}{1.2}
    % \vspace{0.3cm}
    \resizebox{0.90\textwidth}{!}{
    \begin{tabular}{c|cccccccc}
    \hline
      & \textbf{\begin{tabular}[c]{@{}c@{}}Motion\\ smooth.\end{tabular}} & \textbf{\begin{tabular}[c]{@{}c@{}}Dynamic\\ degree\end{tabular}} & \textbf{\begin{tabular}[c]{@{}c@{}}Spatial\\ relation\end{tabular}} & \textbf{Scene} & \textbf{\begin{tabular}[c]{@{}c@{}}Appear.\\ style\end{tabular}} & \textbf{\begin{tabular}[c]{@{}c@{}}Subject\\ consist.\end{tabular}} & \textbf{\begin{tabular}[c]{@{}c@{}}Back.\\ consist.\end{tabular}} & \textbf{Avg.} \\ \shline
        \ding{55}  &  89.10  &  36.29  &  59.78  &  69.01  &  41.32  & 88.98  &  91.33 & 67.97 \\
        KL   &  88.89  &  36.50  &  59.40  &  69.11  &  41.78  & 89.05  & 91.45  & 68.02 \\ \rowcolor[gray]{0.9}
       Wass. & 89.32 & 36.48 & 63.23 & 71.32 & 42.25 & 91.90 & 92.55 & \textbf{69.58} \\ 
     \hline                                
    \end{tabular}
    }

    \label{tab:abla_result_videogen}
    % \vspace{-1.2em}
\end{table*}

\subsubsection{Policy Optimization for Video Generation}
Table~\ref{tab:abla_result_videogen} compares video generation performance under three regularization settings during reinforcement learning: no regularizer, KL Divergence, and Wasserstein Distance. 
The model optimized with Wasserstein Distance achieves the highest Total score of 69.58, exceeding both the non-regularized baseline, 67.97, and the KL-regularized model, 68.02. 
The main gains appear in dynamic and consistency-related metrics, including Motion Smoothness, 89.32, Spatial Relation, 63.23, and Background Consistency, 92.55. 
Although KL regularization performs slightly better on Dynamic Degree, Wasserstein Distance gives stronger results across multiple metrics related to motion and scene coherence. 
This suggests that it better aligns the video generation policy distribution with the target reward distribution during optimization, which is also reflected in the examples in Fig.~\ref{fig:align_showcase}.

\subsection{Further Analysis}
\begin{figure}
  \centering
  \begin{subfigure}{0.48\columnwidth}
      \centering
    \includegraphics[width=\columnwidth]{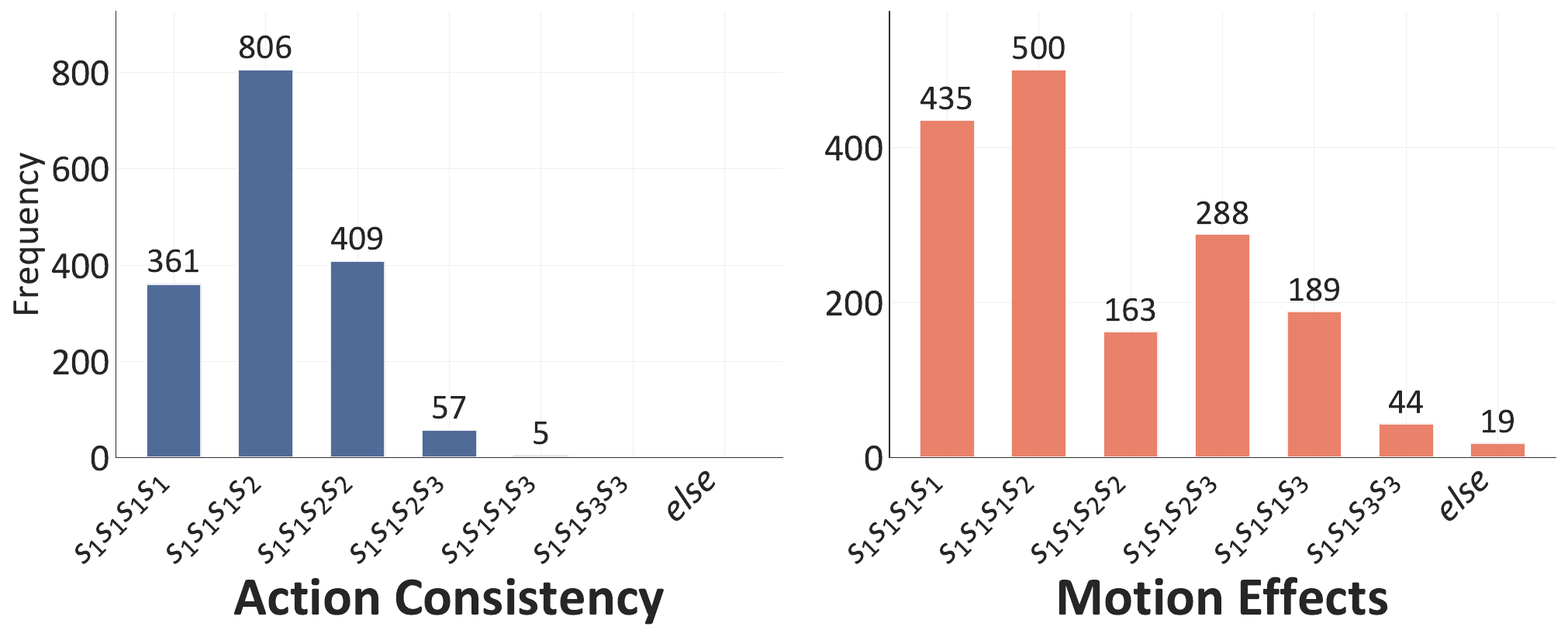}
      \caption{Distribution of samples with different annotator-consistency levels for (left) action and (right) motion.}
      \label{fig:dist_anno_consistency}
  \end{subfigure}
  \hfill
  \begin{subfigure}{0.48\columnwidth}
    \centering
      \includegraphics[width=\columnwidth]{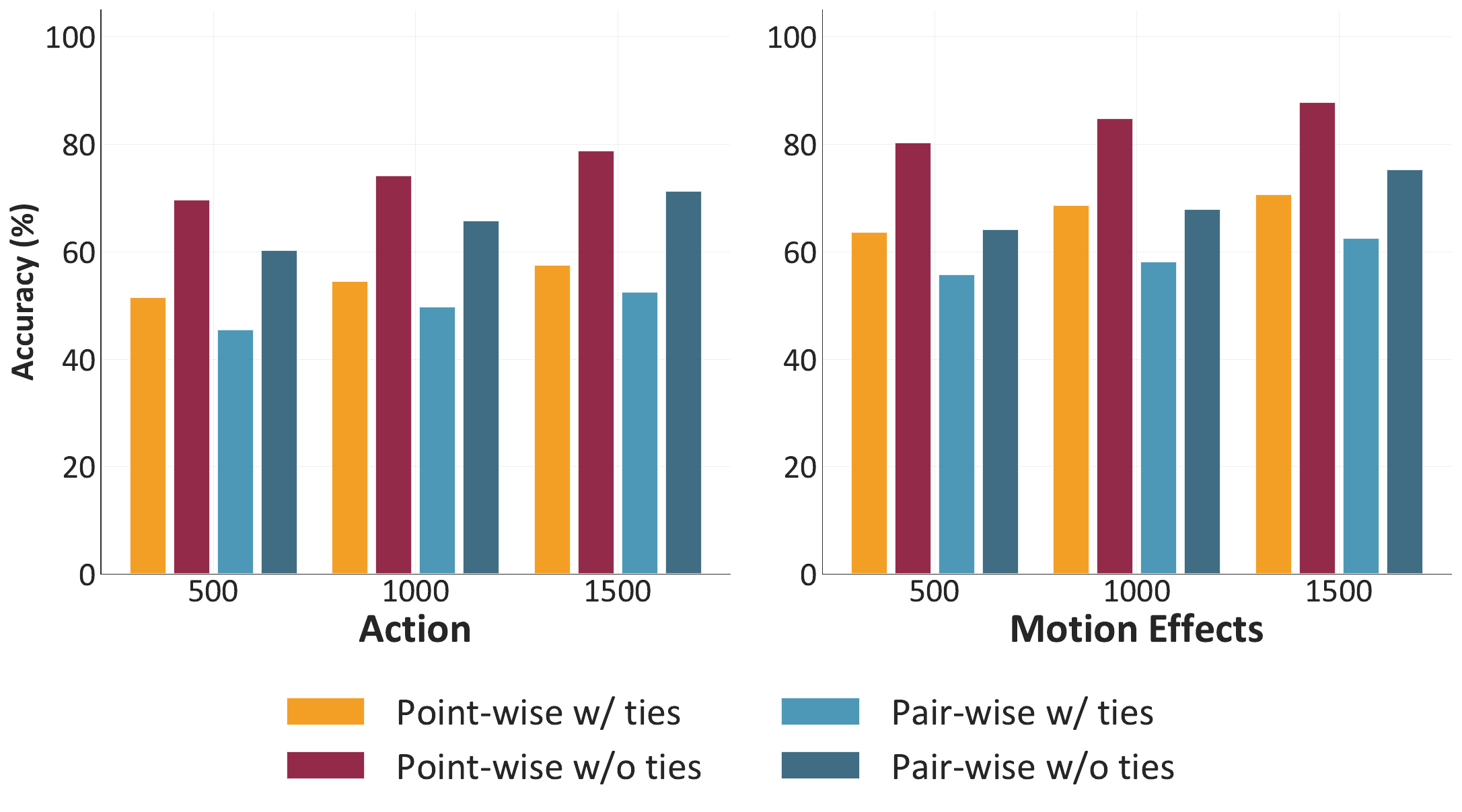}
      \caption{Performance comparison across different data scales shows that under noisy conditions, point-wise training achieves more stable and reliable accuracy than pair-wise training.}
      \label{fig:point_vs_pair}
  \end{subfigure}
  
  \begin{subfigure}{0.48\columnwidth}
    \centering
    \includegraphics[width=\columnwidth]{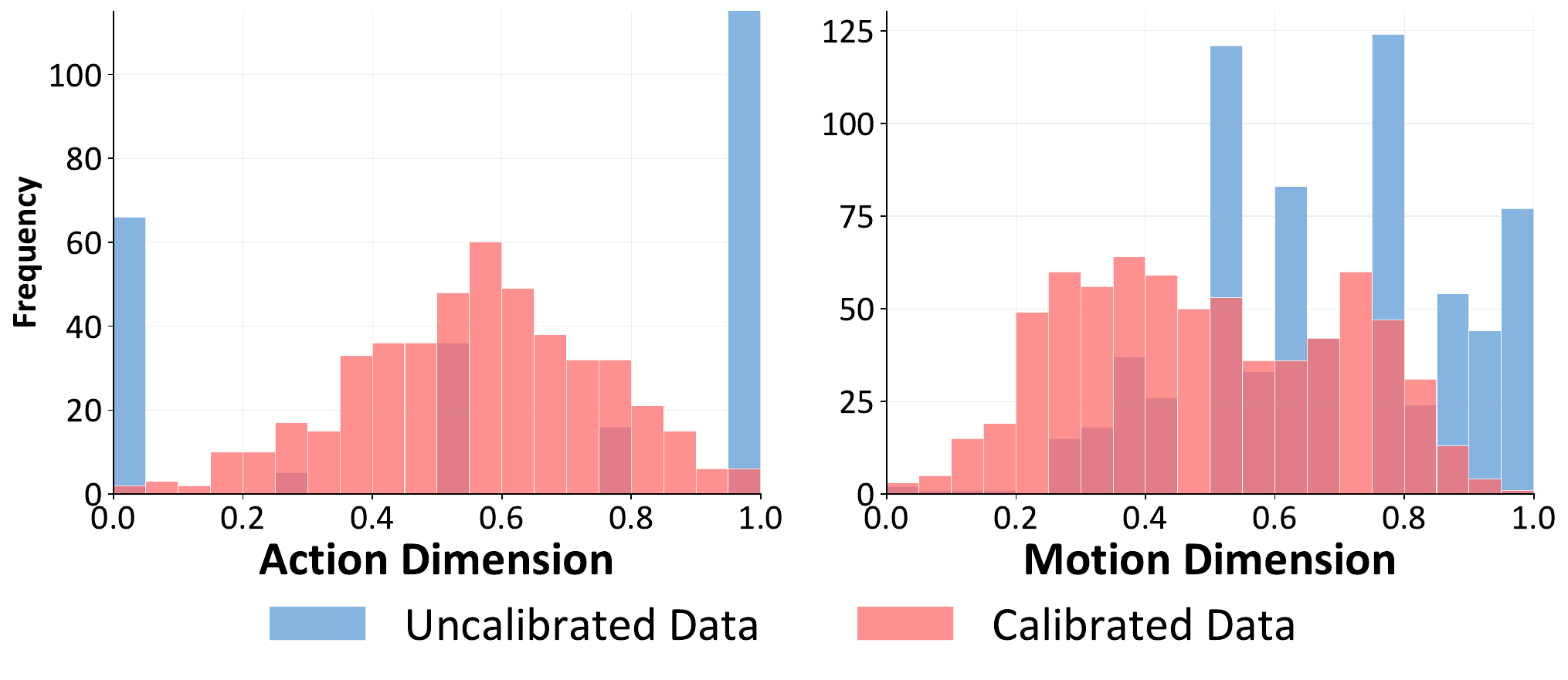}
      \caption{Score distribution before and after Calibration.}
      \label{fig:bef_aft_calibration}
  \end{subfigure}
  \hfill 
  \begin{subfigure}{0.48\columnwidth}
    \centering
    \includegraphics[width=\columnwidth]{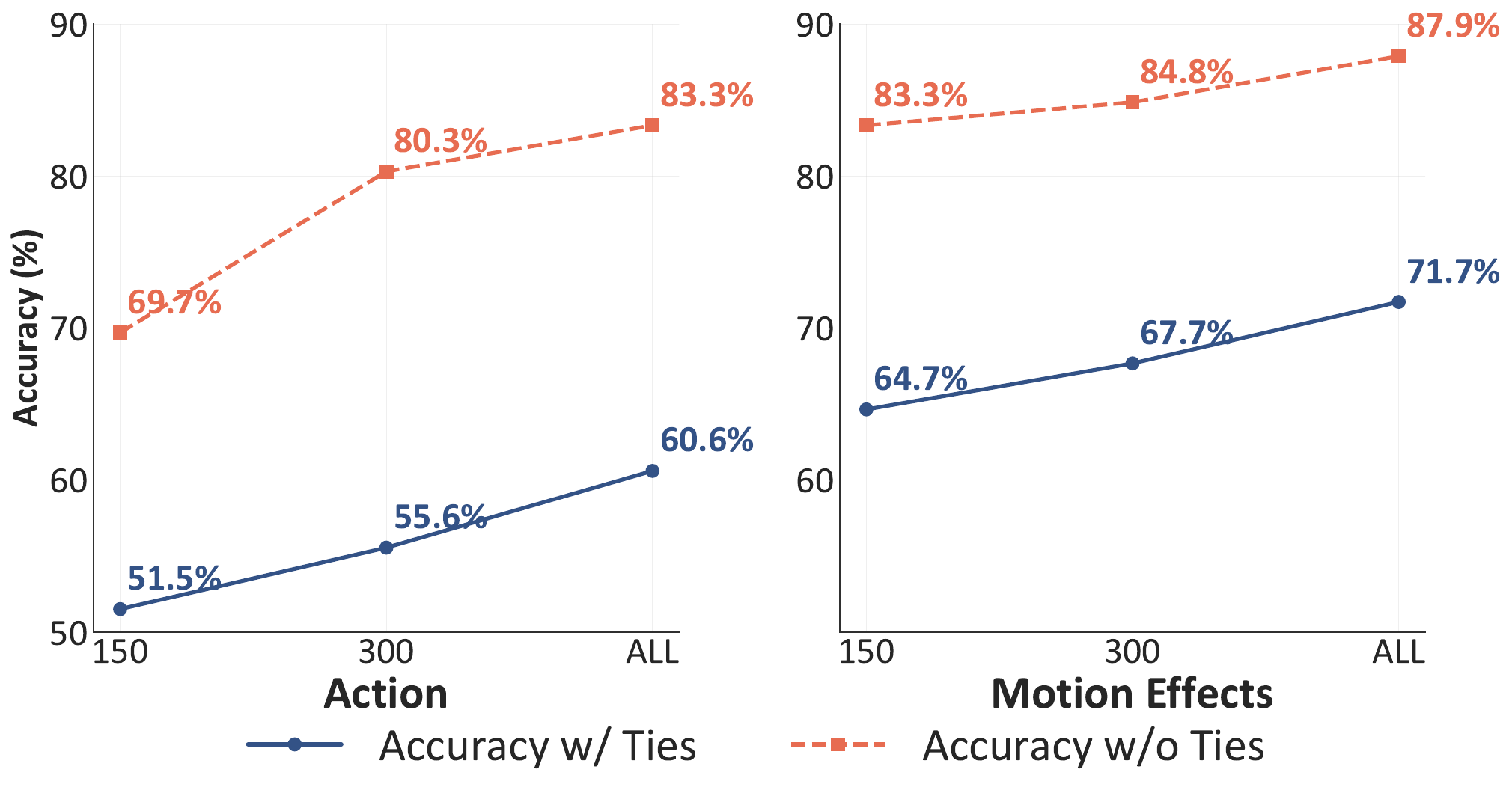}
    \caption{Relationship between the number of calibrated training samples and accuracy on action consistency and motion effects Dimensions.}
    \label{fig:calibrated_scale}

  \end{subfigure}

  \caption{Further analysis of data quality, training robustness, and calibration effects.}
\end{figure}
\subsubsection{Annotation Consistency and Data Quality}
Fig.~\ref{fig:dist_anno_consistency} shows the distribution of sample consistency across different dimensions in the original dataset. 
In the action consistency dimension, besides 1,167 high quality samples, there are 471 low consistency samples; in the Motion Effects dimension, 703 noisy samples are nearly comparable in number to the 935 high quality samples. 
As shown, noisy samples account for a substantial proportion: 28.8\% and 42.9\% of the total samples in the Action and Motion dimensions, respectively.

From empirical evidence, ratings from a single annotator are more susceptible to subjective bias and fail to capture a stable, population-level preference distribution. Even with multiple annotators, a subset of high-complexity samples still induces substantial inter-rater disagreement. 
Such noisy data exhibit high-variance human preferences that do not faithfully reflect genuine, high-quality preference signals, and therefore are unsuitable for model training.

\subsubsection{Paradigm Comparison on Robustness to Noise} In noisy samples, different annotators often assign different scores to the same instance. 
We investigate how this phenomenon influences the choice of training paradigms in Fig.~\ref{fig:point_vs_pair}.
Experimental results show that as the noise level increases, the performance of the pairwise training paradigm degrades substantially.
In such cases, when training reward models with noisy data, the traditional pointwise training approach proves more effective in handling score inconsistencies.
This is because pairwise training relies on accurate pairwise comparisons to optimize preference learning, but in noisy environments, these pairwise relationships can become distorted, hindering the optimization process.
In contrast, the pointwise training method independently optimizes each sample’s score without relying on pairwise comparisons, thereby exhibiting greater stability and robustness with noisy data.
\subsubsection{Visualization on Calibration Effect} 
We analyze the importance of score calibration, as illustrated in Fig.~\ref{fig:bef_aft_calibration}, which shows the changes in score distributions of noisy data before and after calibration.
For the original multi-annotator scores, we use the average rating as the initial score label.
In both the action and motion effects dimensions, we observe high-score samples whose actual quality is poor. 
The action consistency dimension also contains low-score samples with high actual quality. 
Due to the lack of consistency among annotators, it is impractical to assign a unified label to these samples.
With a robust calibration model, we effectively mitigated such scoring inconsistencies.
The calibrated score distributions become smoother and more continuous, substantially reducing the presence of outlier samples in the original distribution.

\subsubsection{Which leads to performance improvement? Quality vs. Quantity.} We investigate how increasing the proportion of calibrated samples affects model performance. 
First, we train the scoring model using absolutely high-quality samples, and then progressively calibrate noisy samples of different scales.
As shown in Fig.~\ref{fig:calibrated_scale}, when the number of calibrated samples increases from 150 to 300 and eventually covers all noisy data, the performance of the reward model trained on the progressively expanded calibrated dataset consistently improves. 
Specifically, in the action consistency dimension, the accuracy (with ties and without ties) increases by 9.1 and 13.6 points, respectively, while in the Motion dimension, it improves by 7.0 and 4.6 points, respectively.

\begin{table}[t]
\centering
\caption{Policy optimization comparison using reward models trained on raw data and calibrated data. RM Type denotes Reward Model Type. We report results for DanceGRPO and FlowGRPO, where each method is evaluated with the original raw reward model and the reward model trained on the calibrated preference distribution.}
\renewcommand{\arraystretch}{1.2}
\resizebox{\textwidth}{!}{%
\begin{tabular}{cccccccccc}
\hline
    \multicolumn{1}{c}{\textbf{Method}} & 
    \multicolumn{1}{c}{\textbf{\begin{tabular}[c]{@{}c@{}}RM\\ Type\end{tabular}}} & 
    \multicolumn{1}{c}{\textbf{\begin{tabular}[c]{@{}c@{}}Motion\\ smooth.\end{tabular}}} & 
    \multicolumn{1}{c}{\textbf{\begin{tabular}[c]{@{}c@{}}Dynamic\\ degree\end{tabular}}} & 
    \multicolumn{1}{c}{\textbf{\begin{tabular}[c]{@{}c@{}}Spatial\\ relation\end{tabular}}} & 
    \multicolumn{1}{c}{\textbf{Scene}} & 
    \multicolumn{1}{c}{\textbf{\begin{tabular}[c]{@{}c@{}}Appear.\\ style\end{tabular}}} & 
    \multicolumn{1}{c}{\textbf{\begin{tabular}[c]{@{}c@{}}Subject\\ consist.\end{tabular}}} & 
    \multicolumn{1}{c}{\textbf{\begin{tabular}[c]{@{}c@{}}Back.\\ consist.\end{tabular}}} & 
    \textbf{Avg.} \\ 
\hline
\multirow{2}{*}{\begin{tabular}[c]{@{}c@{}}Dance\\ GRPO\end{tabular}} & Raw(Full)  & 88.10          & 36.40          & 63.90          & 71.20          & 42.10          & 89.50         & 91.70          & 68.99          \\
 & Calibrated & \textbf{88.30} & \textbf{37.20} & \textbf{65.60} & \textbf{72.10} & \textbf{43.30} & \textbf{90.00} & \textbf{93.30} & \textbf{69.97} \\ 
\hline
\multirow{2}{*}{\begin{tabular}[c]{@{}c@{}}Flow\\ GRPO\end{tabular}} & Raw(Full)  & 89.00          & 38.10          & 64.90          & 71.80          & 42.70          & 90.30         & 92.20          & 69.86          \\
 & Calibrated & \textbf{89.40} & \textbf{39.40} & \textbf{66.20} & \textbf{72.60} & \textbf{43.90} & \textbf{90.90} & \textbf{93.70} & \textbf{70.87} \\ 
\hline
\end{tabular}%
}
\label{tab:baselines}
\end{table}

\subsubsection{Generalization Across Policy Optimizers}  Table~\ref{tab:baselines} evaluates our calibrated distributional reward model with FlowGRPO~\cite{liu2026flow} and DanceGRPO~\cite{xue2025dancegrpo}  on Wan2.1-T2V-1.3B. Replacing the raw reward model with the calibrated one consistently improves both optimizers, increasing the average score respectively. 
These results suggest that our calibration strategy provides a generally useful reward signal rather than an optimizer-specific gain.
\section{Conclusion}
\label{sec:conclusion}

In this paper, we propose a unified preference-aware video generation framework to align video generation with human perceptual judgments of quality. 
Our approach introduces three core mechanisms: the elite-guided filtering for reliable preference data calibration, a multidimensional reward distribution to capture the nature of human preference and the Wasserstein-based distributional alignment integrated into GRPO to enforce alignment with the global structure of human preference. 
Experimental results show that our framework significantly improves the reliability of reward signals and enhances the consistency of human preference in the generated videos. 

% \clearpage  % TODO FINAL: This \clearpage needs to be removed from both review and camera-ready versions.

\section*{Acknowledgements}
The work described in this paper was partially supported by InnoHK initiative, The Government of the HKSAR, and Laboratory for AI-Powered Financial Technologies.

\bibliographystyle{splncs04}
% \bibliography{main}
\bibliography{
    main,
    references/benchmark,
    references/method,
    references/models,
    references/motivation,
    references/rl,
    references/videogen,
    references/reward
}
\end{document}